\documentclass[final]{cvpr}

\usepackage{times}
\usepackage{epsfig}
\usepackage{graphicx}
\usepackage{amsmath}
\usepackage{amssymb}
\usepackage{color, colortbl}
\usepackage[pagebackref=true,breaklinks=true,colorlinks,bookmarks=false]{hyperref}
\definecolor{Gray}{gray}{0.9}
\def\cvprPaperID{****} % *** Enter the CVPR Paper ID here

\def\cvprPaperID{15} % *** Enter the CVPR Paper ID here
\def\confYear{CVPR 2021}

\begin{document}

\title{Compact and Effective Representations for Sketch-based Image Retrieval}  % **** Enter the paper title here
\author{Pablo Torres\\
DCC, University of Chile \\
Av. Beauchef 851, Santiago, Chile\\
{\tt\small pablo.torres.dessi@gmail.com}
\and
Jose M. Saavedra\\
Impresee Inc.\\
600 California St, San Francisco, USA\\
{\tt\small jose.saavedra@impresee.com}
}

\maketitle

\begin{abstract}

Sketch-based image retrieval (SBIR) has undergone an increasing interest in the community of computer vision bringing high impact in real applications. For instance, SBIR brings an increased benefit to eCommerce search engines because it allows users to formulate a query just by drawing what they need to buy. However,  current methods showing high precision in retrieval work in a high dimensional space, which negatively affects aspects like memory consumption and time processing. Although some authors have also proposed compact representations, these drastically degrade the performance in a low dimension. Therefore in this work, we present different results of evaluating methods for producing compact embeddings in the context of sketch-based image retrieval. Our main interest is in strategies aiming to keep the local structure of the original space. The recent unsupervised local-topology preserving dimension reduction method UMAP fits our requirements and shows outstanding performance, improving even the precision achieved by SOTA methods. We evaluate six methods in two different datasets. We use Flickr15K and eCommerce datasets; the latter is another contribution of this work. We show that UMAP allows us to have feature vectors of 16 bytes improving precision by more than $35\%$.
\newline
\newline
\copyright 2021 IEEE
\end{abstract}

%%%%%%%%% BODY TEXT - ENTER YOUR RESPONSE BELOW
\section{Introduction}
\label{sec:intro}
\begin{figure}[ht!]
    \centering
    \includegraphics[width = \linewidth]{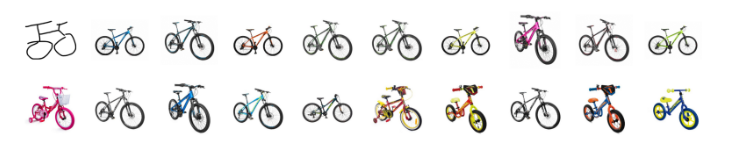}
    \caption{An example of retrieval produced by a sketch-based image retrieval engine in the context of e-commerce. Here, a user draws a small tricycle showing the purchase intention. The first image is the query an the others are the retrieved images.}
    \label{fig:sbir_example}
\end{figure}

Sketch-based image retrieval (SBIR) has undergone an increasing interest in the computer vision community, mainly because it benefits many modern applications. An SBIR method aims to retrieve photos or regular images from a collection, resembling a hand-drawing used as a query.  Figure \ref{fig:sbir_example} shows some results generated by a sketch-based image retrieval engine in the context of e-commerce, where users simply draw what they need to buy.

Indeed, e-commerce is a very attractive application where a SBIR engine brings tremendous advantages, especially at a time when online shopping is becoming more and more popular. The success of a e-commerce depends on the effectiveness of its search engine, that allows buyers to find what they are looking for. Querying by sketching is effective, easy and enjoyable. It is effective because a sketch can semantically express the buyer's desire; it is also easy because users can draw directly on a screen; and it is enjoyable because users feel drawing as a game.

We have seen a particular interest in the computer vision community on image retrieval guided by sketch queries, during the last decade. At the beginning of this period, researchers focused on low-level features, proposing variations of the histogram of orientations to deal with sketch-like images \cite{Saavedra:2014, Eitz:2011, Hu:2013}. Other researchers also used  mid-level features \cite{Saavedra:2013:J, Saavedra:2015} to represent sketches, particularly detecting primitive shapes, called \emph{keyshapes}, on the images. However, the explosion of deep-learning,  which showed outperforming effectiveness in diverse computer vision tasks, rerouted the SBIR research toward these models that significantly increase the underlying efficacy \cite{Sangkloy:2016, Bui:2018}. At the end, the architectures based on siamese backbones with triplet loss, trained in an incremental manner showed the best performance on different SBIR datasets \cite{Bui:2018}.% More recently, a sketch representation generated from a transformer-based model was proposed by Ribeiro et al. [11] with competitive results.

The effectiveness achieved by SOTA methods on sketch-based retrieval has made it possible to move quickly from the scientific context to industrial applications. However, applications like an eCommerce search engine also place new challenges beyond the effectiveness itself. In this vein, searching time and memory consumption are two aspects that SBIR methods should consider. One way to address this challenge is by proposing methods that produce embeddings in a low-dimension feature space.

Commonly, SBIR methods produce floating-point representations in high-dimension feature space (p.ej. 1024). This fact has a big impact on the resources required by a company offering a search engine service since all the feature vectors of a store catalog need to be loaded into the memory to allow online querying. To deal with this problems some researches have proposed compact representations  \cite{paper:liu_2017, article:bui_2017}. However, these methods still show low effectiveness in retrieval.

%$Some approaches have been proposed to produce compact representations [], even producing a binary embedding [], but degrading significantly the effectiveness measure by the precision on retrieval. \cite{article:bui_2017}
A simple manner to obtain a reduced feature space is applying a dimensional reduction method. Dimension reduction techniques fall in two categories, those that seek to preserve the global structure of the data those that prefer to preserve the local structure \cite{paper:mcinnes_2020}. The latter category brings more benefits for image retrieval, as the techniques falling in that category try to extract features that better represent the neighborhood of each point (keeping the local structure). This behaviour can also increase the retrieval process's precision, bringing back other relevant objects that in the original spaces fell out of its corresponding neighborhood.

Therefore, in this work we are interested in generating low-dimensional embeddings in the context of sketch-based image retrieval, and improving the retrieval precision showed by SOTA methods. To this end, we leverage the local structure showed by the method proposed by Bui et al. \cite{Bui:2018}, and reduce the space dimension trough an unsupervised reduction technique that focus in keeping the local topology of the original space. In this vein, we show that UMAP (Uniform Manifold Approximation and Projection for
Dimension Reduction) \cite{paper:mcinnes_2020}  fits perfectly with our objective. We conduct diverse experiments comparing binary reduction, PCA, UMAP and even adding a reduction layer to the baseline model.

We also propose a new evaluation dataset that reflects the variations of images we can find in eCommerce applications. Besides, to compare our proposal with current methods, we use the public dataset Flickr15K \cite{Hu:2013}.

Our results indicate that we can obtain very small representations (16 ~ 32 bytes) using UMAP,  improving even the retrieval precision. Here, using a 16-bytes embedding, we get a mAP of $57\%$ in Flicr15k dataset and $19\%$ in the eCommerce dataset.

%The good thing about reducing keeping local topology is that we can increar the precision on retrieval. Our experiments show SOTA results, around 56$\%$ in FLickr 15K , using 32 bytes for each embedding. In the ecommerce dataset we also observe an increment in precision (22$\%$), when we reduce to 32 bytes.

This document is organized as follows. In Section \ref{sec:related_work} we describes the related work. Section \ref{sec:proposal} is devoted to describe our approach in detail. Section \ref{sec:experiments} describe the evaluation experiments, and finally Section \ref{sec:conclusions} presents the conclusions.
 
%Contributions 
%- experiments with reduced representations
%- test on real datasets PEPEGANGA will be 

\section{Related Work}
\label{sec:related_work}
During the last period, we have seen significant advances in sketch-based image retrieval. In the beginning, researchers focused on low-level features, mainly using some variants of the histogram of orientations \cite{Saavedra:2012, Hu:2013, Eitz:2011}. Researchers also proposed methods based on mid-level representations \cite{Saavedra:2015, Saavedra:2012}, where sketch-like images are represented through a distribution of primitives called \emph{keyshapes}. However, the explosion of deep learning has shown outperforming results in computer vision tasks, allowing us to move toward more effective methods in the context of sketch-based retrieval.

In this context, the architectures that have shown the best results on similarity search are those combining siamese nets with triplet loss \cite{Yu:2016, Bui:2018, Sangkloy:2016}, especially when a fine-grained search is desired.  Among these methods, the work of Bui et al. \cite{Bui:2018} attracts our interest because it achieves high precision on public datasets on sketch-based image retrieval. This method proposes a 4-stage incremental methodology for training a network capable of producing a feature space where sketches and photos can exist together. The four stages are designed in such a way that they can incrementally improve their discriminatory power. To this end, they also use siamese and triplet networks jointly with a cross-entropy loss but trained from a coarse-grained similarity at the beginning to a fine-grained similarity at the end. 

Although SBIR methods have evolved positively, achieving high precision in different public datasets, their application in real contexts has still been limited by the underlying feature space's size. Indeed, methods showing high precision require hundreds of floating-point values per image, which is prohibited for catalogs with millions of images.

Regarding the previous problem, some researchers have also been focused on reducing the high-dimension space in sketch-based retrieval and image retrieval, in general. However, the proposed reduction algorithm rapidly degrades the effectiveness of retrieving. 
%For example, methods like DTSH \cite{paper:liu_2016} and DSH \cite{paper:wang_2016}  are focused on obtaining a binary feature space for image retrieval.

The simplest way to obtain a binary embedding is by adding an activation function at the network's head so that the final output can be thresholded to get a binary value. Examples of these functions are \emph{sigmoid} and \emph{tanh}. For instance, we can use a threshold equal to 0.5 in the case of the \emph{sigmoid} function. This method has the advantage of being easy to incorporate into an existing network. % although there is low control in the thresholding process. % the level of thresholding other than the chosen function.

Deep Supervised Hashing (DSH) \cite{paper:liu_2016}  and Deep Triplet Supervised Hashing (DTSH) \cite{paper:wang_2016}  are focused on obtaining a binary feature space for image retrieval.  DSH uses a network with three convolutional-pooling layers and two fully connected layers. The method forces the output to be binary by the following pairwise loss function:
\begin{equation*}
\label{eq:DTSH_loss}
\begin{aligned}
\mathcal{L}(b_{1},b_{2}) & =\frac{1}{2}(1-y)||b_{1}-b_{2}||^2_2 \\
& + \frac{1}{2}y\max(0,m-||b_{1}-b_{2}||^2_2 )\\
& +\lambda(||\ |b_{1}|-\mathbf{1}||_1+||\ |b_{2}|-\mathbf{1}||_1)
\end{aligned}
\end{equation*}

where the first two terms work as the contrastive loss, and the third term is the binarization term that forces the outputs to be binary. Here,  $m$ is a margin parameter, $y=1$ if the $b_{1},b_{2}$ are similar and 0 otherwise, and $\lambda$ is a weighting parameter that controls the strength of the binarization term. Once trained, the binary codes can be obtained by applying the sign function.

Deep Triplet Supervised Hashing (DTSH) \cite{paper:wang_2016} is another method for getting binary representations. DTSH loss comes from the likelihood of producing good triplets, where the similarity between pairs is computed as the inner product between the corresponding feature vectors. The  loss is defined as follows: 

%the negative log triplet label likelihood and a different regularizer term for the binarization that read as follows:

\begin{equation*}
\label{eq:DSH_loss}
\begin{aligned}
\mathcal{L}(u_i,u_j,u_k) & = \log(1+e^{\Theta_{i,j}-\Theta_{i,k}-\alpha}) - (\Theta_{i,j}-\Theta_{i,k}-\alpha)\\
& +\lambda(||b_{i}-u_i||_2^2+||b_{j}-u_j||_2^2+||b_{k}-u_k||_2^2)
\end{aligned}
\end{equation*}

where $\Theta_{i,j}=\frac{1}{2}u_i^Tu_j$, $b_i=sgn(u_i)$, $\alpha$ is the margin parameter, and $\lambda$ is a weighting parameter that controls the strength of the regularizer. Similar to DSH, the third term is devoted to forcing binary outputs.

The use of reduction techniques over the embeddings generated by a model is another alternative to produce compact embeddings. For instance, Bui et al.\cite{article:bui_2017}  uses Product Quantization and PCA to get small feature vectors. However, the precision achieved by the methods is still low.

Dimension reduction techniques can be divided into two categories: those that try to preserve the original space's global structure and those that are more focused on keeping the space's local structure. PCA is the most usual method of the former category; it shows competitive results for small reduction ratios, but its effectiveness drastically decreases when the reduction ratio is high. For instance, in the context of sketch-based retrieval, a reduction from  2048 dimensions to 8 produces a degradation of about $400\%$ in mAP.

In the second category, t-SNE \cite{article:vanDerMaaten_2008} and UMAP \cite{paper:mcinnes_2020} are methods showing outperforming results.  Contrary to PCA, these methods take into account local information of each point in the original space. This behavior allows the method to extract relevant features for each point, which improves the precision of an image retrieval method. Among this category, UMAP has shown to outperform t-SNE,  as it preserves more of the global structure with superior run time performance. Moreover, UMAP  is able to scale to larger dataset sizes.

%is a method devoted to small target dimensions (e.g 2-3), what make it appropriately for visualization purposes only.  In contrast, UMAP y a more general reduction technique wiht lower reduction time.

Therefore, in this work, we present an evaluation study of compact reduction techniques in the context of sketch-based image retrieval. We compare binary embeddings and embedding generated by dimension reduction techniques as  PCA and UMAP. Our results indicate the superiority of UMAP over the rest of the techniques. Using UMAP allows us to reduce the features space and increase the precision of retrieval. We evaluate these methods on two datasets, one well known by the community and the other related to a real application in eCommerce. In this manner, we can get results closer to real applications. 

Our results show that we can reduce to 16 bytes improving precision in both datasets. We observe a precision gain over $35\%$ in both datasets, passing from 0.42 to 0.57 in Flickr15K and from 0.14 to 0.19 on eCommerce dataset. In both datasets the reduction ratio allows to optimize the used memory as well as time searching significantly, that is critical for real environments.
%TODO: Mencionar map achieved in Pepepganga y Flickr15K

%------------------------------------------------------------------------
\section{Compact Representation Methods}
\label{sec:proposal}
%here we need to describe our base network
In this section, we describe the methods we evaluate to produce compact representations in the context of sketch-based image retrieval. We conduct the experiments through two datasets. The first one, Flick15k, is a public dataset commonly used in this context; and the other, named eCommerce,  is closer to real applications like eCommerce searching. The latter dataset is another contribution of this work,  and we will make it public shortly.  

All the methods we discuss in this section use a convolutional neural network either as a backbone on which a reduction layer is added or to produce an initial feature space. To this end, we implemented a baseline model based on the work of Bui et al. \cite{Bui:2018}. As our interest is on evaluating the impact of compact representation methods, we are not so concerned with replicating the exact results of Bui's work, but rather with replicating the methodology followed by them, which allows us to have effectiveness close to the state-of-the-art.

\begin{figure}
\includegraphics[width = \linewidth]{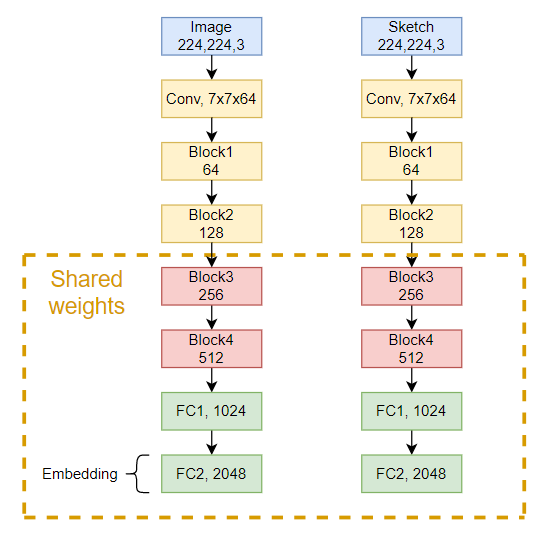}
\caption{Baseline architecture used in this work.}
\label{fig:convnet_scheme}
\end{figure}

%TODO: describe the current architecutre [PABLO]
The baseline architecture is a siamese network composed of two backbones. Each backbone is a ResNet-50 \cite{paper:xie_2016} equipped with two fully-connected layers at the head, where the layer \textbf{FC2} is responsible for generating the feature vector given an input image or sketch. This baseline is trained incrementally, from training both backbones independently to a training guided by triplets $(s,i_p,i_n)$, where $s$ is the anchor sketch, $i_p$ is a positive image with respect to $s$, and $i_n$ is a negative image.  Similar to the work of Bui et al. \cite{Bui:2018}, we use Flikcr25K, and Sketchy \cite{Sangkloy:2016} as training datasets. In our implementation, the dimension of the feature spaces produced by the baseline is 2048.  Figure \ref{fig:convnet_scheme} depicts an scheme our baseline architecture.

In the following lines, we describe the methods we implement to produce compact representations.
%method
\subsection{In-situ Reduction Layer (RL)}
We can produce small representations adding a new fully-connected layer (\textbf{FC3}) at the end of the baseline and allowing the model to learn a transformation from  FC2's outputs to get a smaller embedding. The new layer FC3 should be understood as a reduction layer. 

\subsection{Binary Sigmoid Layer (BSL)}
The most straightforward manner to obtain a binary representation is by adding a sigmoid function at the network's head. Therefore, output values greater than 0.5 will be labeled with 1 and the others with 0. We implemented this approach, adding a sigmoid function after the layer FC2.

\subsection{Deep Supervised Hashing (DSH)}
In the original paper \cite{paper:liu_2016}, DSH uses a small network to learn binary representations. We believe that by using a deeper network, as our baseline, the retrieval precision should be maintained or even improved. Thus, we incorporate a fully connected layer at the end of our baseline whose size represents the target dimension. Finally, we adapt the DSH loss defined in Eq. \ref{eq:DSH_loss} to work with triplets in the following manner:

\begin{equation*}
\label{eq:new_DSH_loss}
\begin{aligned}
\mathcal{L_{DSH}}(b_{q},b_{p},b_n) & = \\
& \frac{1}{2}\max(0,m-||b_{q}-b_{n}||^2_2+ ||b_{q}-b_{p}||^2_2) \\
& +\lambda(||\ |b_{q}|-\mathbf{1}||_1  \\
& + ||\ |b_{p}|-\mathbf{1}||_1+||\ |b_{n}|-\mathbf{1}||_1)
\end{aligned}
\end{equation*}
where  $b_q$, $b_p$, and $b_n$ are the embeddings produced by the network, where the first one is a sketch query and the others are  the embedding of the positive and negative images, respectively.

In addition, we add a cross-entropy loss to leverages the image labels. To this end, we add a classification layer after the reduction layer. To combine the adapted DSH loss ($\mathcal{L}_{DSH}$) and the cross-entropy loss ($\mathcal{L}_{CE}$), we use a weighting parameter $\alpha = 0.4$. The final loss is defines as follows:

\begin{equation*}
\mathcal{L} =  \alpha \frac{\mathcal{L_{DSH}}}{D} + (1 - \alpha)  \mathcal{L_{CE}},
\end{equation*}
where $D$ is the size of the target feature space.

%TODO PABLO
\subsection{Deep Triplet Supervised Hashing (DTSH)}
As before, we add a fully connected layer at the end of the baseline network to get the reduced output. As DTSH was originally proposed to work with triplets, we do not need to modify it in order to be plugged into our network. However, analogously to the DSH approach, the original loss function is combined with the cross-entropy loss. To this end, we add a fully-connected layer for classification at the end of the network producing a final loss as a weighted sum of both losses. We also normalize the DTSH loss dividing it by the size of the target feature space.
%TODO PABLO
\subsection{Principal Component Analysis (PCA)}
For the sake of completeness, we incorporate PCA as a method for dimension reduction.  This also  provides a comparison baseline in the context of dimension reduction techniques. We test the goodness of PCA with different target sizes, and compare its performance against a local-structure preserving method.
%TODO PABLO
\subsection{Uniform Manifold Approximation and Projection (UMAP)}
UMAP is an unsupervised technique for dimension reduction that leverages the local structure of the points in the original space to learn a compact manifold. Keeping the local structure of an original space is an important property for image retrieval methods.

When a reduction method tries to maintain each point's neighborhood during the reduction process, it indirectly finds relevant features. If the original space forms clusters around each point, the dimension reduction technique will choose features related to those clusters since it tries to preserve the space's local topology. Consequently, the reduced space will pull in points to their corresponding neighborhood. UMAP \cite{paper:mcinnes_2020} was developed with this property in mind, where the local structure of each point is defined based on its closest neighbors.

UMAP requires two parameters, the size of the neighborhood ($K$) and the minimum distance apart (min\_dist) that points are allowed to be in the low dimensional representation. Small $\textrm{min\_dist}$ will create more compact clusters. In all our experiments we use $K = 15$, and $\textrm{min\_dist} = 0.1$.

%The parameters used with UMAP were a neighborhood of 15 elements and a minimum distance of $0.1$. The neighborhood size controls how UMAP balances local versus global structure in the data, while the minimum distance controls how tightly UMAP is allowed to pack elements together.

\section{Experimental Evaluation}
\label{sec:experiments}
In this section, we describe all the experiments we run to evaluate the goodness of the methods described in Section \ref{sec:proposal}. We will describe datasets, the training process, and the metrics used during the experiments.

\subsection{Datasets Description}
\begin{enumerate}
\item \textbf{Flickr15K}: This public dataset is commonly used for testing sketch-based retrieval methods. It has a catalog of 14,660 regular images (photos) divided into 33 classes related to
historical places, objects, animals, plants, etc. Besides, it has 330 hand-drawn sketches for querying (10 for each class).
\item \textbf{eCommerce}: We propose this dataset to have a better understanding of the behavior of SBIR methods in real applications like eCommerce. We choose this context because it represents one of the most modern applications of sketch-based retrieval with massive use.

This dataset is composed of a catalog with 10600 photos related to products we can find in a eCommerce. These images are distributed over 133 classes according to their shape. The dataset has a diversity of products such as household appliances, clothing, toys, etc. Figure \ref{fig:ecommerce_catalog} depicts a sample of photos we can find in this dataset. Also, we include 665 hand-drawn sketches, 5 per class. To collect the sketches, we ask people to make a drawing resembling a picture we pick randomly. Figure \ref{fig:ecommerce_queries} show a sample of sketch queries we can find in the eCommerce dataset.

%took inspiration from images of the dataset, but trying to make the drawings as general as possible with respect to the classes. A sample of the images in the catalos is depicted in Figure \ref{fig:ecommerce_dataset}. In addition, a sample of queries of this dataset is showed in Figure \ref{fig:ecommerce_queries}.
\begin{figure}[ht!]
    \centering
    \includegraphics[width = \linewidth]{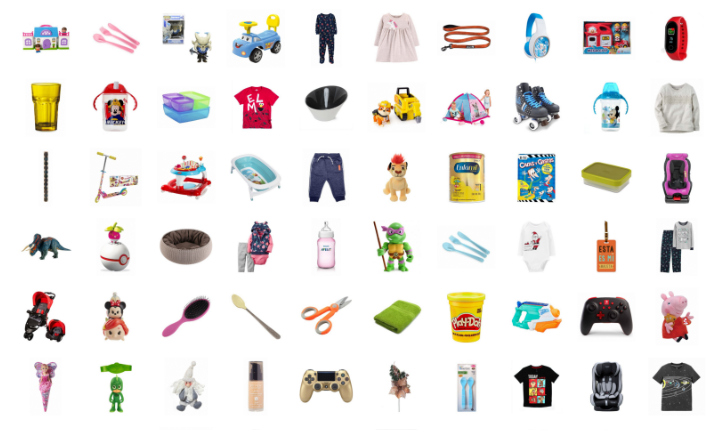}
    \caption{A sample of images in the eCommerce catalog.}
    \label{fig:ecommerce_catalog}
\end{figure}
\begin{figure}[ht!]
    \centering
    \includegraphics[width = \linewidth]{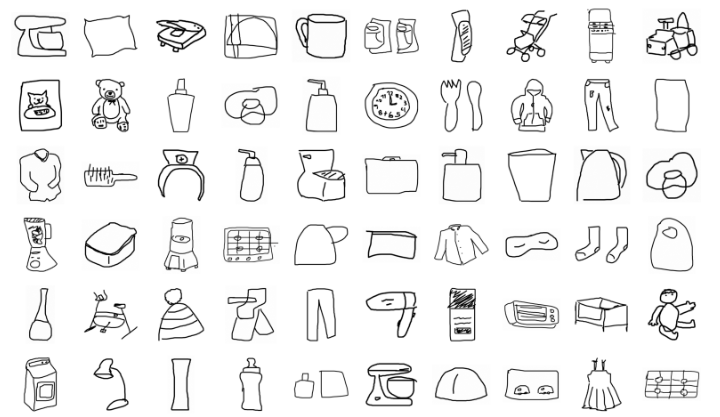}
    \caption{An example of sketch queries in the eCommerce catalog.}
    \label{fig:ecommerce_queries}
\end{figure}
\end{enumerate}

\subsection{Training Description}
To train the reduction methods based on neural networks (RL, BSL, DSH, DTSH), we extend the baseline with the modifications discussed above, loading the corresponding weights and training them for 10 epochs using the Sketchy dataset \cite{Sangkloy:2016}. This dataset provides a set of sketches with their corresponding positive photos. Thus, to form the triplets required for training the reduction models, we use the already labeled positive pairs. Then, for each pair we add a negative photo in an online manner. To this end, we randomly pick, as negative, any image different from those labeled as positive one with respect to the underlying sketch. The training was carried out for each target dimension.

In the case of using the dimension reduction techniques (PCA and UMAP) we conduct a different strategy. Since these are unsupervised methods, the transformation they can learn should be inferred with data following the same distribution of the testing set.  For this purpose,  we apply cross-validation over the Flickr15K and eCommerce datasets. We divide the image set into three parts, using two partitions for fitting the transformation and the remaining one for testing.  It is important to mention, that only the photos of the datasets were used for the training process, not the sketches. The measurements generated by these techniques are computed as an average value after running the cross-validation procedure.
%For the time processing metrics, we carry out the training and evaluations on the complete datasets to be consistent with the rest of the methods
\subsection{Metrics}
In this evaluation we consider four metrics :
\begin{itemize}
    \item \textbf{mAP}: This measures the average precision on retrieving each relevant image for each query. However,  when a query has many relevant objects, the mAP decreases drastically as the last retrieved relevant objects appear far away from the front of the ranking. This fact does not reflect a real situation, like in the context of eCommerce search engines, where the precision should be guided by the first relevant objects. Thus, to have a better approximation of the performance, we include other metrics. These are Mean Reciprocal Rank and Recall-Precision. In addition, we also measure the search time with different dimensions.
    \item \textbf{Mean Reciprocal Rank}: This measures the precision of the first retrieved relevant object.
    \item \textbf{Recall-Precision}: This shows the performance through a 2D graphic that crosses the recall with precision. This allows us to have a global view of the performance as it depicts the precision achieved by a method at different levels of recall. In real applications, we are more interested in seeing the precision for the first relevant objects (e.g. precision at recall 0.2).
    \item \textbf{Time processing} : We measure the searching time for different dimension reduction sizes using UMAP and PCA.
\end{itemize}

\subsection{Result Analysis}
\subsubsection{Baseline Performance}
We implemented the architecture defined in Section following the methodology proposed in the work of Bui \cite{Bui:2018}. We tested this baselines model in Flickr15K achieving a mAP of 0.42. Although the precision is not the same as reported in the original paper, we consider this is not a critical problem as we are interested in measuring the impact of compact reductions against a SOTA baseline. In this work our implementation of the Bui's work is considered as the baseline for all the experiments. Table \ref{tab:baseline} shows the mAP achieved by of the baseline in Flickr15K and eCommerce datasets.

\begin{table}
\centering
\begin{tabular}{l|c|c}
& \textbf{Flickr15K} & \textbf{eCommerce} \\\hline
Baseline & 0.421 & 0.145
\end{tabular}
\caption{Mean Average Precision achieved by the baseline in Flickr15 and eCommerce dataset.}
\label{tab:baseline}
\end{table}

\subsubsection{Precision}
In this section, we report the mAP achieved by the following methods: Reduction Layer (RL), Binary Sigmoid Layer (BSL), Deep Triplet Supervised Hashing (DTSH), Deep Hashing (DSH), Principal Component Analysis (PCA) and Uniform Manifold Approximation and Projection (UMAP), which are described in Section \ref{sec:proposal}.

\begin{figure}[ht!]
    \centering
    \includegraphics[width = \linewidth]{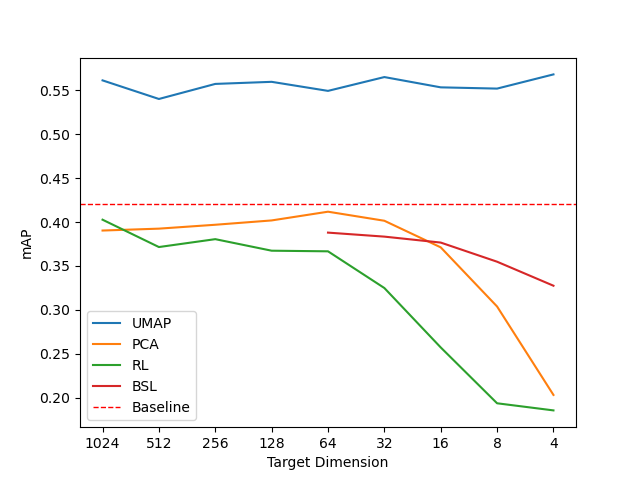}
    \caption{mAP achieved by the compact methods for different target dimensions in  Flickr15K dataset.}
    \label{fig:methods_map}
\end{figure}

Table \ref{tab:map_flickr15k} reports the mAP achieved by the proposed methods in the Flickr15K, and Table \ref{tab:map_ecommerce} presents the results in the eCommerce dataset. In both cases, UMAP beats the other methods by a large margin,  improving even the baseline's performance. Although we present results for different target dimensions, the main interest of this work is in analyzing  the behavior of strategies on low dimensions. In this case, we observe the good behavior achieved by UMAP for 8 or 4 dimensions, which allows us to represent a feature vector with 32 bytes and 16 bytes, respectively.  Figure \ref{fig:methods_map} illustrates the behavior of the compact methods for different target dimensions in the Flickr15K dataset.

In both datasets, UMAP allows us to drastically reduce the size of the feature vectors, from 2048 to 4 dimensions (16 bytes), and increase the precision up to $35\%$ with respect to the baseline. UMAP achieves a mAP of 0.57 in Flickr15K and 0.19 in eCommerce dataset. The preservation of the local topology in the original space plays a key role in this achievement.

\begin{figure}[ht!]
    \centering
    \includegraphics[width = \linewidth]{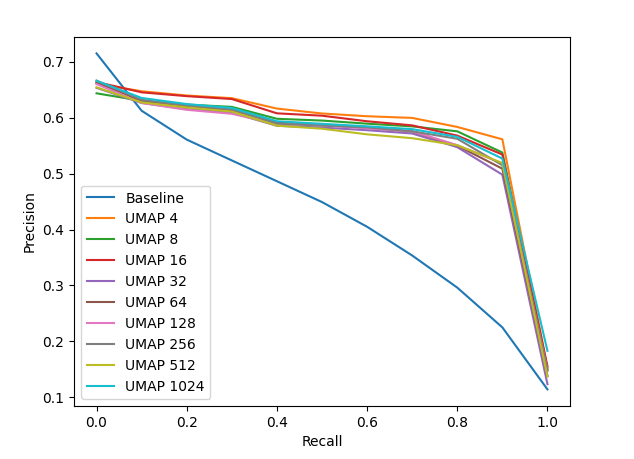}
    \caption{Recall-Precision chart for Flickr15K dataset.}
    \label{fig:rp_flickr15k}
\end{figure}

\begin{figure}[ht!]
    \centering
    \includegraphics[width = \linewidth]{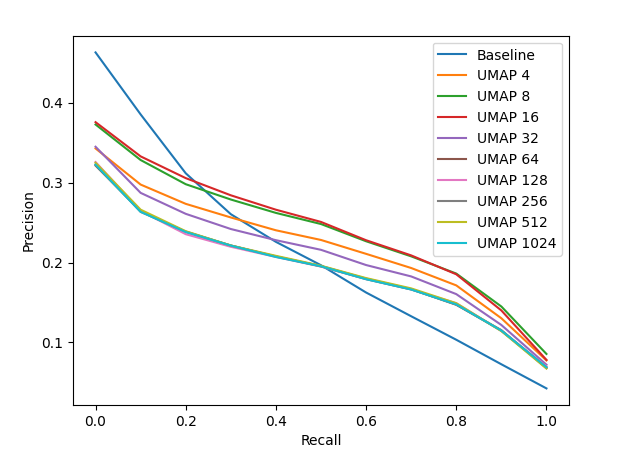}
    \caption{Recall-Precision chart for the eCommerce dataset.}
    \label{fig:rp_ecommerce}
\end{figure}

In addition, we can assess the precision of the different methods for different values of recall through the Recall-Precision chart. Figures \ref{fig:rp_flickr15k} and \ref{fig:rp_ecommerce}  depict the relation of recall and precision of UMAP for different target dimensions in Flickr15K and eCommerce datasets, respectively. In these charts we observe that for low-recall values, the precision achieved by the baseline is better, and when we walk to greater recall UMAP behaves better. This is the effect of extracting relevant features of local structures. In this way, objects that felt apart from their corresponding neighborhood in the original space, are attracted to them in the reduced space.

We can also observe that UMAP produces better results in Flickr15K than in eCommerce dataset. This happens because the baseline's precision is higher in the former dataset, which indicates the presence of a better local-topology. Indeed, better the local-structure better the features that UMAP can extract.

\newcolumntype{g}{>{\columncolor{Gray}}c}
\begin{table}
\centering
\begin{tabular}{l|c|c|c|c|c|g}
\textbf{Size} & \textbf{RL} & \textbf{BSL}  & \textbf{DTSH} & \textbf{DSH} & \textbf{PCA} & \textbf{UMAP} \\\hline
1024 & 0.40 & - & - & - & 0.39 & 0.56 \\ 
512 & 0.37 & - & - & - & 0.39 & 0.54 \\
256 & 0.38 & - & - & - & 0.40 & 0.56\\
128 & 0.37 & - & - & - & 0.40 & 0.56\\
64 & 0.37 & 0.39 & 0.19 & 0.38 & 0.41 & 0.55\\
32 & 0.32 & 0.38 & 0.22 & 0.34 & 0.40 & 0.57\\
16 & 0.26 & 0.38 & 0.28 & 0.30 & 0.37 & 0.55\\
8 & 0.19 & 0.35 & 0.31 & 0.26 & 0.30 & 0.55\\
4 & 0.18 & 0.33 & 0.33 & 0.25 & 0.20 & \textbf{0.57}
\end{tabular}
\caption{Mean Average Precision achieved by different compact representation methods in the Flickr15K dataset. The first column is the feature space's dimension.}
\label{tab:map_flickr15k}
\end{table}

\begin{table}
\centering
\begin{tabular}{l|c|c|c|c|c|g}
\textbf{Size} & \textbf{RL} & \textbf{BSL}  & \textbf{DTSH} & \textbf{DSH} & \textbf{PCA} & \textbf{UMAP} \\\hline
1024 & 0.13 & - & - & - & 0.15 & 0.15 \\ 
512 & 0.12 & - & - & - & 0.15 & 0.16 \\
256 & 0.12 & - & - & - & 0.15 & 0.16\\
128 & 0.12 & - & - & - & 0.15 & 0.16\\
64 & 0.10 & 0.11  & 0.05 & 0.13 & 0.14 & 0.16\\
32 & 0.09 & 0.11  & 0.06 & 0.12 & 0.13 & 0.18\\
16 & 0.06 & 0.09 & 0.08 & 0.10 & 0.09 & 0.21\\
8 & 0.5 & 0.07 & 0.10 & 0.08 & 0.11 & \textbf{0.21}\\
4 & 0.03 & 0.05 & 0.10 & 0.06 & 0.06 & 0.19
\end{tabular}
\caption{Mean Average Precision achieved by different compact representation methods in the eCommerce dataset. The first column is the feature space's dimension.}
\label{tab:map_ecommerce}
\end{table}

To complement this study, we include Tables \ref{tab:mrr_fickr15k} and \ref{tab:mrr_ecommerce} that reports the MRR (mean reciprocal rank) for each method in  Flickr15K and eCommerce dataset, respectively. The results are coherent with what we see in the Recall-Precision charts. None of the compact methods beats the baseline's performance. The baseline achieves an MRR of 0.714 in Flickr15K and 0.40 in the eCommerce dataset. Using a representation of 16 bytes (4 dimensions), UMAP achieves 0.60 in Flick15K and 0.26 in the eCommerce dataset. Even though UMAP experiments a degradation of performance for low-level recall, it is still better than the other competitors.

\begin{table}
\centering
\begin{tabular}{l|c|c|c|c|c|g}
& RL & BSL  & DTSH & DSH & PCA & UMAP \\\hline
1024 & 0.68 & - & - & - & 0.62 & 0.59\\ 
512 & 0.62 & - & - & - & 0.62 & 0.60\\
256 & 0.63 & - & - & - & 0.62 & 0.62\\
128 & 0.60 & - & - & - & 0.61 & 0.60\\
64  & 0.62 & 0.66 & 0.34 & 0.67 & 0.61 & 0.60\\
32  & 0.56 & 0.64 & 0.39 & 063 & 0.59 & 0.61\\
16  & 0.45 & 0.62 & 0.45 & 0.60 & 0.54 & 0.59\\
8   & 0.33 & 0.59 & 0.47 & 0.55  & 0.45 & 0.59\\
4   & 0.30 & 0.55 & 0.45 & 0.53 & 0.29 & 0.60
\end{tabular}
\caption{MRR of evaluated methods in Flickr15K dataset, where MRR of baseline is 0.714.}
\label{tab:mrr_fickr15k}
\end{table}

\begin{table}
\centering
\begin{tabular}{l|c|c|c|c|c|g}
& RL & BSL  & DTSH & DSH & PCA & UMAP \\\hline
1024 & 0.38 & - & - & - & 0.39 & 0.21 \\ 
512 & 0.36 & - & - & - & 0.38 & 0.21 \\
256 & 0.36 & - & - & - & 0.38 & 0.20\\
128 & 0.34 & - & - & - & 0.38 & 0.21\\
64 & 0.30 & 0.32 & 0.19 & 0.40 & 0.37 & 0.21\\
32 & 0.27 & 0.34 & 0.19 & 0.37 & 0.32 & 0.23\\
16 & 0.17 & 0.30 & 0.21 & 0.34 & 0.26 & 0.28\\
8 & 0.13 & 0.27  & 0.23 & 0.27 & 0.23 & 0.29\\
4 & 0.09 & 0.20 & 0.21 & 0.24 & 0.10 & 0.26
\end{tabular}
\caption{MRR of evaluated methods in eCommerce dataset, where MRR of baseline is 0.40.}
\label{tab:mrr_ecommerce}
\end{table}

\subsubsection{UMAP Drawbacks}
\begin{figure}[ht!]
    \centering
    \includegraphics[width = \linewidth]{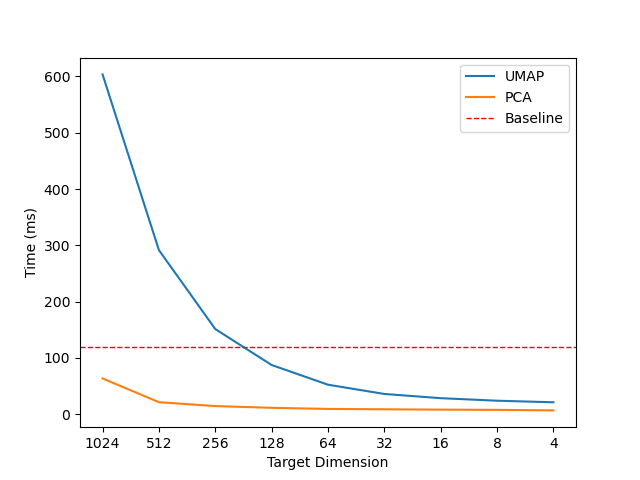}
    \caption{Searching time for different target dimensions in the Flickr15K dataset. This includes query time transformation.}
    \label{fig:time_umap}
\end{figure}

The most important disadvantage of using UMAP is its non-parametric nature. This brings an extra cost for transforming the query into a reduced space. However, as we can see in Figure \ref{fig:time_umap}, when we are in low-dimension space, lower than 16 dimensions, the transformation cost is negligible. 

\begin{figure}[ht!]
    \centering
    \includegraphics[width = \linewidth]{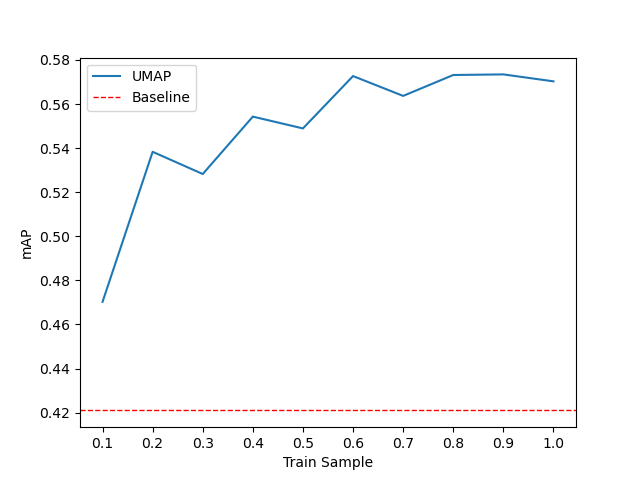}
    \caption{mAP achieved by UMAP in Flickr15K training UMAP with different portions of the complete dataset.}
    \label{fig:portions}
\end{figure}

The non-parametric property of UMAP requires an additional space to store the mapping between the original space and the target space. It could limit its applicability to real contexts. However, we observe that to understand de local-structure of the original space, UMAP requires only a portion of the complete catalog. How much data UMAP requires depends on the variability of the data set. For instance, in the Flickr15K dataset, UMAP beats the baseline using up to $10\%$ of the data. Figure \ref{fig:portions} illustrates the performance of UMAP using different portions of the Flickr15K dataset for training.

Finally, in Table \ref{tab:sota}, we also compare our proposal versus published results in the Flickr15K dataset. We note that our proposal based on UMAP improves the precision using  very a low dimension, 16 bytes, which provides great impact in real applications.

\begin{table}[]
    \centering
    \begin{tabular}{l|c|c}
         Method &  Size (bytes) & mAP \\\hline
         Mulitregression \cite{Bui:2018} &  1024 & 0.53 \\
         Compact Representation \cite{article:bui_2017}  &  7 & 0.22  \\
         Compact Representation \cite{article:bui_2017}  &  400 & 0.25  \\
        \rowcolor{Gray}     
         Ours (UMAP) & 16 & 0.57
    \end{tabular}
    \caption{mAP achieved by state-of-the-art methods on sketch-based image retrieval. }
    \label{tab:sota}
\end{table}

\subsubsection{Qualitative Results}
\begin{figure}[ht!]
    \centering
    \includegraphics[width = \linewidth]{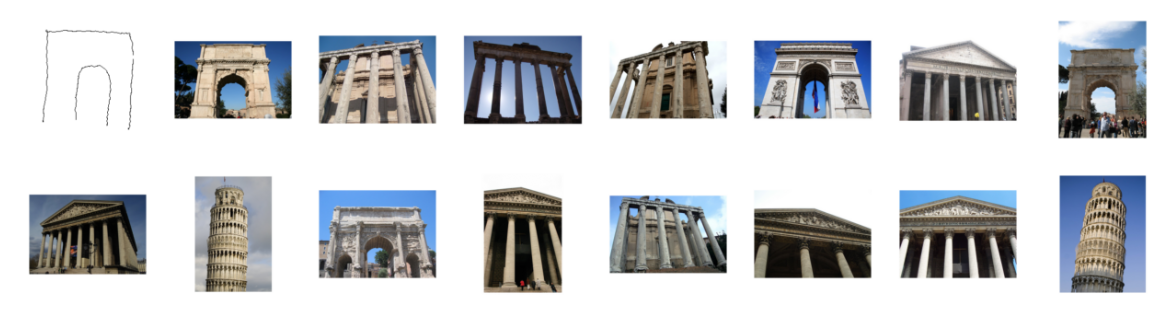}
    \caption{An example of retrieval in the Flickr15K dataset with the baseline method.}
    \label{fig:sample_flickr_original}
\end{figure}

\begin{figure}[ht!]
    \centering
    \includegraphics[width = \linewidth]{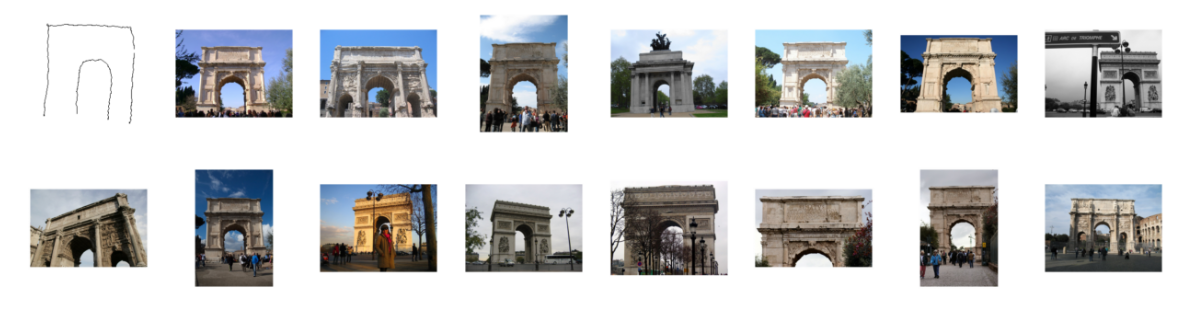}
    \caption{An example of retrieval in the Flickr15K dataset with the UMAP method.}
    \label{fig:sample_flickr_umap}
\end{figure}

\begin{figure}[ht!]
    \centering
    \includegraphics[width = \linewidth]{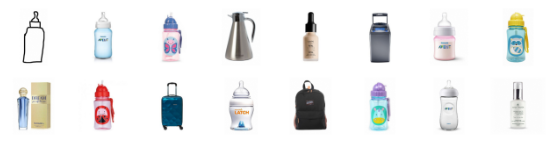}
    \caption{An example of retrieval in the eCommerce dataset with the original method.}
    \label{fig:sample_ecommerce_original}
\end{figure}

\begin{figure}[ht!]
    \centering
    \includegraphics[width = \linewidth]{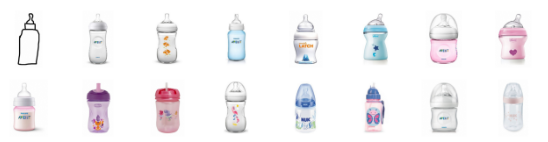}
    \caption{An example of retrieval in the eCommerce dataset with the UMAP method.}
    \label{fig:sample_ecommerce_umap}
\end{figure}

To visually understand the goodness of the UMAP reduction technique, we add qualitative results.  Figures \ref{fig:sample_flickr_original} and  \ref{fig:sample_flickr_umap} show example of retrieval in the Flickr15K dataset, with the baseline method and UMAP, respectively. While Figures \ref{fig:sample_ecommerce_original} and  \ref{fig:sample_ecommerce_umap} show example of retrieval in the eCommerce dataset, with the baseline method and UMAP, respectively. We observe that in both cases, UMAP increases the precision after retrieving the first 15 results. Figure \ref{fig:sample_flickr_umap} shows the effectiveness of the UMAP based method to retrieve more images of \emph{Triumphal arch}, while  Figure \ref{fig:sample_ecommerce_umap} shows  how the same method retrieves more baby bottles. 


\section{Conclusions}
In this work, we present an evaluation study of different methods to get compact representations in the context of sketch-based image retrieval. We also present a new dataset that represents a real application. This dataset is related to the eCommerce environment, where querying by sketches brings great opportunities. Our experimental results show the superior performance of UMAP, allowing us to work with feature vectors of 16 bytes and increasing the precision achieved by a SOTA baseline in $35\%$ in both datasets, Flickr15K and the one proposed in this work.  
\label{sec:conclusions}

{\small
\bibliographystyle{ieee_fullname}
\bibliography{egbib}

\begin{thebibliography}{10}\itemsep=-1pt

\bibitem{article:bui_2017}
T. Bui, L. Ribeiro, M. Ponti, and J. Collomosse.
\newblock Compact descriptors for sketch-based image retrieval using a triplet
  loss convolutional neural network.
\newblock {\em Computer Vision and Image Understanding}, 164:27--37, 2017.

\bibitem{Bui:2018}
Tu Bui, Leonardo Ribeiro, Moacir Ponti, and John Collomosse.
\newblock Sketching out the details: Sketch-based image retrieval using
  convolutional neural networks with multi-stage regression.
\newblock {\em Computers \& Graphics}, 71:77--87, 2018.

\bibitem{Eitz:2011}
Mathias Eitz, James Hays, and Marc Alexa.
\newblock How do humans sketch objects?
\newblock {\em ACM Trans. Graph. (Proc. SIGGRAPH)}, 31(4):44:1--44:10, 2012.

\bibitem{Hu:2013}
Rui Hu and John Collomosse.
\newblock A performance evaluation of gradient field hog descriptor for sketch
  based image retrieval.
\newblock {\em Computer Vision and Image Understanding}, 117(7):790--806, July
  2013.

\bibitem{paper:liu_2016}
H. {Liu}, R. {Wang}, S. {Shan}, and X. {Chen}.
\newblock Deep supervised hashing for fast image retrieval.
\newblock In {\em 2016 IEEE Conference on Computer Vision and Pattern
  Recognition (CVPR)}, pages 2064--2072, 2016.

\bibitem{paper:liu_2017}
Li Liu, Fumin Shen, Yuming Shen, Xianglong Liu, and Ling Shao.
\newblock Deep sketch hashing: Fast free-hand sketch-based image retrieval.
\newblock In {\em 2017 {IEEE} Conference on Computer Vision and Pattern
  Recognition, {CVPR} 2017, Honolulu, HI, USA, July 21-26, 2017}, pages
  2298--2307. {IEEE} Computer Society, 2017.

\bibitem{paper:mcinnes_2020}
Leland McInnes, John Healy, and James Melville.
\newblock Umap: Uniform manifold approximation and projection for dimension
  reduction, 2020.

\bibitem{Saavedra:2014}
J.~M. {Saavedra}.
\newblock Sketch based image retrieval using a soft computation of the
  histogram of edge local orientations (s-helo).
\newblock In {\em 2014 IEEE International Conference on Image Processing
  (ICIP)}, pages 2998--3002, 2014.

\bibitem{Saavedra:2015}
Jose~M. Saavedra and Juan~Manuel Barrios.
\newblock Sketch based image retrieval using learned keyshapes {(LKS)}.
\newblock In {\em Proceedings of the British Machine Vision Conference 2015,
  {BMVC} 2015, Swansea, UK, September 7-10, 2015}, pages 164.1--164.11, 2015.

\bibitem{Saavedra:2013:J}
Jose~M. Saavedra and Benjamin Bustos.
\newblock Sketch-based image retrieval using keyshapes.
\newblock {\em Multimedia Tools and Applications}, 2013.

\bibitem{Saavedra:2012}
Jose~M. Saavedra, Benjamin Bustos, Tobias Schreck, Sang~Min Yoon, and
  Maximiliam Scherer.
\newblock {Sketch-based 3D Model Retrieval using Keyshapes for Global and Local
  Representation}.
\newblock In {\em Eurographics Workshop on 3D Object Retrieval}, pages 47--50,
  2012.

\bibitem{Sangkloy:2016}
Patsorn Sangkloy, Nathan Burnell, Cusuh Ham, and James Hays.
\newblock The sketchy database: Learning to retrieve badly drawn bunnies.
\newblock {\em ACM Transactions on Graphics (proceedings of SIGGRAPH)}, 2016.

\bibitem{article:vanDerMaaten_2008}
Laurens van~der Maaten and Geoffrey Hinton.
\newblock Visualizing data using {t-SNE}.
\newblock {\em Journal of Machine Learning Research}, 9:2579--2605, 2008.

\bibitem{paper:wang_2016}
Xiaofang Wang, Yi Shi, and Kris~M. Kitani.
\newblock Deep supervised hashing with triplet labels, 2016.

\bibitem{paper:xie_2016}
Saining Xie, Ross~B. Girshick, Piotr Doll{\'{a}}r, Zhuowen Tu, and Kaiming He.
\newblock Aggregated residual transformations for deep neural networks.
\newblock {\em CoRR}, abs/1611.05431, 2016.

\bibitem{Yu:2016}
Q. {Yu}, F. {Liu}, Y. {Song}, T. {Xiang}, T.~M. {Hospedales}, and C.~C. {Loy}.
\newblock Sketch me that shoe.
\newblock In {\em 2016 IEEE Conference on Computer Vision and Pattern
  Recognition (CVPR)}, pages 799--807, 2016.

\end{thebibliography}
}

\end{document}